\documentclass{article}
\usepackage[utf8]{inputenc}

\title{Improving  NER's Performance with Massive financial corpus}
\author{Han zhang}
\date{July 2020}

\usepackage[square,sort,comma,numbers]{natbib}
\usepackage{graphicx}

\begin{document}

\maketitle
\section{ABSTRACT}
Training large deep neural networks needs massive high quality annotation data, but the time and labor costs are too expensive for small business. We start a company-name recognition task with a small scale and low quality training data, then using skills to enhanced model training speed and predicting performance with minimum labor cost. The methods we use involve pre-training a lite language model such as Albert-small or Electra-small in financial corpus, knowledge of distillation and multi-stage learning. The result is that we raised the recall rate by nearly 20 points and get 4 times as fast as BERT-CRF model.
\section{INTRODUCTION}
Named Entity Recognition (NER) is one of the basic tasks in natural language processing. The main task of NER is to identify and classify proper names such as names of people, places, meaningful quantitative phrases, and date in the text \citep{Chinchor1997}. Named entity recognition technology is an essential part of many natural language processing technologies such as information extraction\citep{PetkovaProximity}, information retrieval\citep{Guo2009Named}, machine translation\citep{Lee2017Improving} and question answering system \citep{Moll2006Named}. From the whole process of language analysis, named entity recognition belongs to the category of unregistered word recognition in lexical analysis. Named entities are the most numerous unlogged words, the most difficult to identify and the most influential to the effect of word segmentation. According to the evaluation results of SIGHAN(http://www.sighan.org/) Bakeoff data, the loss of segmentation precision caused by unregistered words is at least 5 times greater than the ambiguity, indicating the importance of the status of named entities.

We generally divide Named Entities into two categories: generic Named Entities (e.g., person and location) and domain-specific Named Entities (e.g., enterprise, product, and financing institution)\citep{Jing2020A}. In this paper, we mainly focus on financial NEs in chinese financial news. Company name recognition is the most common task in entity recognition in the financial field, which makes sense to assess the risks of financial institutions. In order to dig out the negative information of enterprises from the massive financial news, the first step is to identify the enterprise name mentioned in the news. For risk assessment, the omission of any company's negative information may cause irreparable losses, so model's recall rate is the most important indicator. The recognition speed is also important because the huge amount of online news are updated in real time. So our goal is to recognize company names in the financial news at high speed and high recall rate. 

With the birth of large pre-training language model such as BERT\citep{Jacob2018BERT}, the classic combination of pre-training model and CRF has become the most common NER model in the industry. However, due to the huge magnitude of BERT's parameters, the disadvantages such as slow prediction speed and easy overfitting have also attracted much attention from the academic community. In the last two years, various lightweight versions of BERT have been coming out, such as DistillBERT\citep{Sanh2019DistilBERT}, FastBERT\citep{Liu2020FastBERT}, DistillBERT\citep{Sanh2019DistilBERT}, Albert\citep{Lan2019ALBERT}, Electra\citep{Clark2020ELECTRA} and so on. They have the similar performance as BERT-base, but the training and prediction speed have been improved several times. Two of the most influential models are Albert and Electra, both open sourced by Google. 

Despite having excellent models, the lack of annotated data is a problem faced by many algorithmic engineers. Some researchers try to study a strategy to improve the training effect in the case of sparse training data. J. Foley et al \citep{Foley2018Named} propose exploring named entity recognition as a search task, where the named entity class of interest is a query, and entities of that class are the relevant “documents”. They presents an exploration of CRF-based NER models with handcrafted features and of how to transform them into search queries. L. Chen et al\citep{liang2020bond} propose a BERT-based two-stage training algorithm under distant supervision, which avoids using large amounts of manual annotation data.

From two aspects of data enhancement and learning strategy, we propose a model tuning strategy suitable for the financial field with sparse training data. Our main work is divided into three parts:
\begin{enumerate}
\item Use financial corpus for pre-training to enhance the training of downstream tasks.
\item Designed a multi-stage learning strategy to improve the learning effect.
\item A knowledge distillation strategy was adopted to enhance the small model.
\end{enumerate}
Our code and part of datasets are available at https://github.com/Hanlard/
Electra\_CRF\_NER.
\section{RELATED WORK}
The main research directions of named entity recognition are: rule-based and lexicographical methods, statistics-based methods and deep learning-based methods. 
\subsection{Rule and dictionary based approaches}
Rules-based methods mostly use the rule templates constructed by linguistics experts and the selected features including statistical information, punctuation marks, keywords, indicators and direction words, position words, and center words. Such systems mostly depend on the knowledge base and dictionary, and use pattern and string matching as the main methods. Rule and dictionary based methods are the earliest ones used in named entity recognition\citep{Xiang2005Chinese}.

Generally speaking, the performance of rule based methods is better than statistics-based methods when extracted rules can accurately reflect language phenomena. However, these rules often depend on the specific language, domain and text style, and the compilation process is time-consuming and difficult to cover all the language phenomena, which is especially prone to errors and the system portability is not good, so linguistics experts are required to rewrite the rules for different systems. Another disadvantage of the rule-based approach is that the cost is too high, the system construction cycle is long, the portability is poor, and the knowledge base of different fields needs to be established to improve the system recognition ability. The rule-based NER approach is more suitable for specific fields where the context is simple and data is scarce\citep{Eftimov2017A}.
\subsection{Statistics-based approach}
Methods based on statistical machine learning mainly include: hidden markov model\citep{Branimir2010Context}, maximum entropy\citep{Jung2012Maximum}, support vector machine\citep{Mansouri2008A}, conditional random field\citep{Sobhana2011Conditional}, etc. 

Among the four learning methods, the maximum entropy model has a compact structure and good universality. Its main disadvantage is that the training time is very complex, and sometimes it even leads to unbearable training cost. In addition, due to the need for explicit normalization calculation, the cost is relatively high. Conditional random field provides a characteristic flexible and globally optimal labeling framework for named entity recognition, but it also has the problems of slow convergence speed and long training time. In general, the maximum entropy and support vector machines are higher in accuracy than the hidden markov model, but the hidden Markov model is faster in training and recognition due to the higher efficiency of viterbi algorithm in solving the named entity class sequence. Hidden markov model is more suitable for some real-time requirements such as short text named entity recognition.

The method based on statistics has a high requirement for feature selection\citep{Pan2012Research}, so it is necessary to select various features that have an impact on the task from the text and add these features into the feature vector. Considering the main difficulties and characteristics of a given named entity, consider selecting a feature set that effectively reflects the characteristics of that entity.The main approach is to extract the features from the training corpus through the statistics and analysis of the language information contained in the training corpus.The related features can be divided into specific word features, context features, dictionary and part of speech features, stop-word features, core word features and semantic features, etc.
Statistics-based methods also rely heavily on corpora, while there are few large-scale general corpora that can be used to construct and evaluate named entity recognition systems.
\subsection{Deep Learning-based approach}
In recent years, DL-based NER models become dominant and achieve state-of-the-art results. Compared to feature based approaches, deep learning is beneficial in discovering hidden features automatically. The main advantages of deep learning are as follows\citep{Jing2020A}: 
\begin{enumerate}
\item Powerful vector representation. Deep learning saves significant effort on designing NER features. The traditional feature-based approaches require considerable amount of engineering skill and domain expertise. Deep learning models, on the other hand, are effective in automatically learning useful representations and underlying factors from raw data. 
\item The powerful computing power and nonlinear mapping capability of neural network. Compared with linear models (e.g., loglinear HMM and linear chain CRF), deep-learning models are able to learn complex and intricate features from data via non-linear activation functions.
\item Deep neural NER models can be trained in an end-to-end paradigm, by gradient descent. Benefits from the non-linear transformation, which generates non-linear mappings from input to output. This property enables us to design possibly complex NER systems.
\end{enumerate}

NER's deep learning research is divided into several branches, including sequence-based method such as LSTM\citep{Zhang2018Chinese}, graph neural network based method such as GCN\citep{Golnar2018GraphNER} , and this year's popular method based on pre-training language model such as BERT\citep{Zi-Niu2019Chinese} et al. Recently, the character-word lattice structure has been proved to be effective for Chinese named entity recognition (NER) by incorporating the word information\citep{li-etal-2020-flat}. 
\subsection{Financial named entity recognition}
In the financial field, the research direction of entity recognition is also extensive. Wang and Xu et al. present a CRF-based method to recognize named entities on financial news texts in three steps which develops CRF feature templates with dictionaries and expanded manual information. They achieve 0.91 precision and 0.92 recall on a chinese financial dataset\citep{Wang2014F}. Zheng et al. develop a rule-based approach that recognizes financial institutions and FI names with dict-based and rank-based methods, highlights the benefits and limitations of specialized versus general purpose approaches.\citep{Zheng2016E}. Zhang et al. study the financial NER and disambiguation problem. They introduce a clearer formulation of the disambiguation problem as an ensemble of methods, each designed to address different sub-problems and challenges. \citep{Zhang2014Entity}
\section{METHODS}
We use pre-training mode (such as BERT-base) and CRF as model framework and use no auxiliary features. For the reason of speed limit, we removed the BiLstm layer between pre-training model and CRF, and the BERT is replaced by a lite version such as Albert-small or Electra-small. The structure of our model can be referred to Figure 1. We take BERT(base)-CRF model as base line and improve speed and performance by the following methods: 1) Pre-training Albert or Electra model on financial corpus, 2) Adopt the multi-stage learning strategy, 3) Adopt the knowledge distillation strategy
\begin{figure}[h!]
\centering
\includegraphics[scale=0.5]{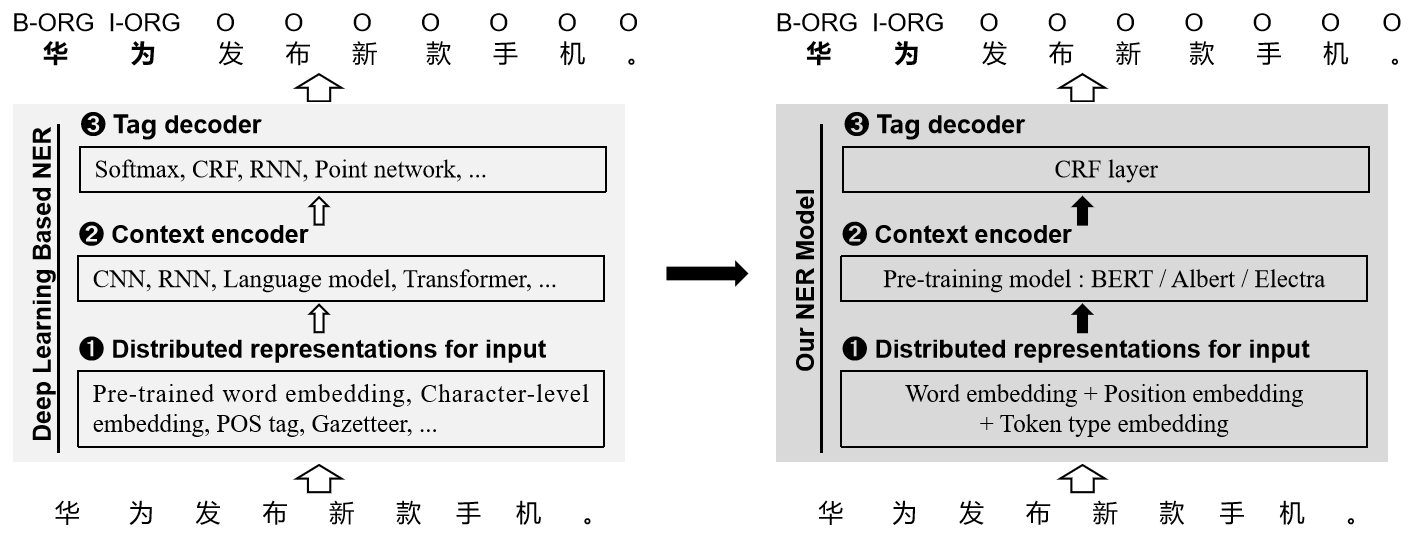}
\caption{The model structure. Left is general deep learning structure for NER, right is the structure for our model }
\label{fig:model}
\end{figure}
\subsection{Pre-training Albert or Electra model on financial corpus}
We use the 1.62 million financial news obtained by web crawler as training corpus to go on pre-training on the basis of the Chinese Electra or Albert model released by the Joint Laboratory of HIT and iFLYTEK Research (HFL, url=https://github.com/ymcui/Chinese-ELECTRA).  Experiments show that pre-training on financial news is helpful to improve the accuracy and training speed of downstream tasks. The detailed results are shown in Figure 2. 
\begin{figure}[h!]
\centering
\includegraphics[scale=0.4]{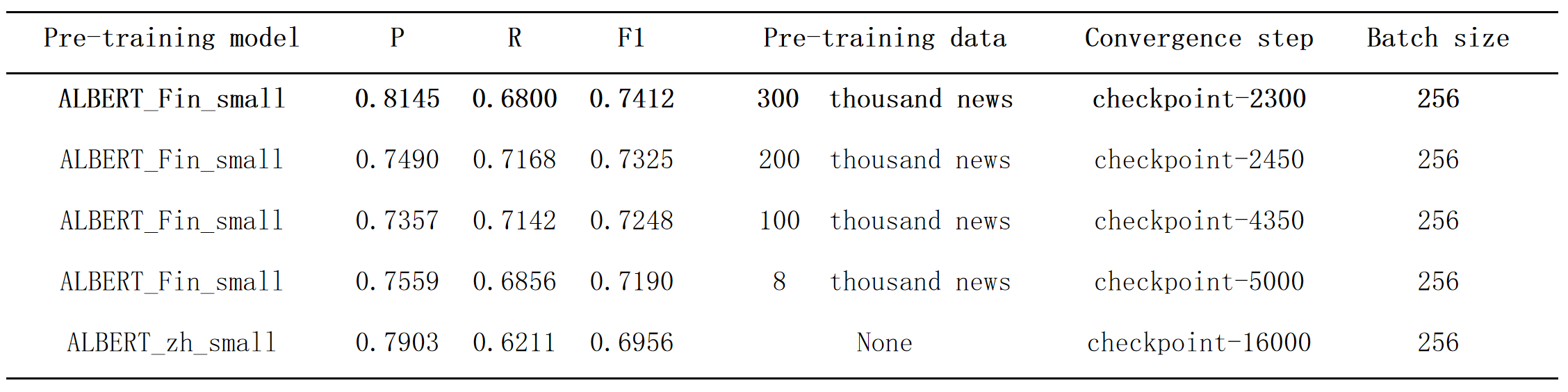}
\caption{Effects of different amounts of pre-training data on downstream tasks}
\label{fig:PreTraining}
\end{figure}
\subsection{Multi-stage learning strategy}
We do trie-tree matching on 350 thousand financial sentences with a lexicon including about 1 million company names. Then we use the chinese NLP tool FOOLNLTK to recogonize some of the omissive company names. So we get the coarse training dataset - "35wSents". The "multi-stage learning strategy" is to train Albert(small)-CRF model on the coarse dataset in the first stage and stop training before the model get over-fitting. We call the first stage "outline-learning". This model after "outline-learning stage" is used to predict the coarse dataset "35wSents". The parts that are not consistent with the prediction and annotation are picked out for manually annotating, which is named "Albert65kError" dataset. We use Albert(small) instead of BERT or Electra, because it contains fewer parameters with the similar model structure and is not easy to overfit. In the second phase, which we call "detail-learning", we fine-tune the model on "Albert65kError" with a small learning rate. Experiments show that multi-stage learning brings an increase of two percentage points, and the two datasets are applicable to other models such as BERT or Electra. The detailed results are shown in Figure 3.
\begin{figure}[h!]
\centering
\includegraphics[scale=0.38]{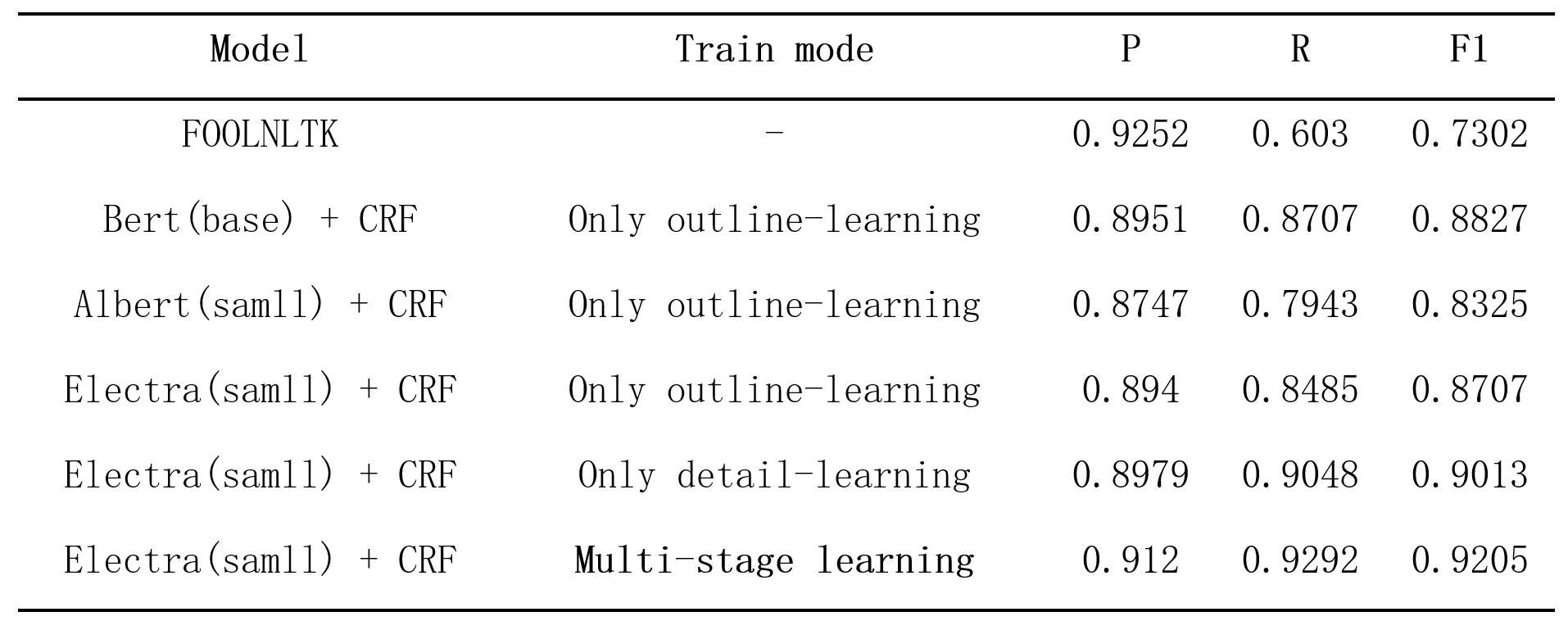}
\caption{Effects of multi-stage learning strategy}
\label{fig:MutiStageLearning}
\end{figure}
\subsection{Knowledge distillation}
Although the BERT-base has better performance, it cannot meet the requirement for speed. Therefore, we adopt a knowledge distillation strategy. Frist we adopt the multi-stage learning strategy to train a Electra(base)-CRF model. Then we use the model to make predictions on a large number of financial news, and use the predicted results as annotated data. Finally, the machine-annotated data is fed to Electra(small)-CRF model for training. Experiments show that Electra(Small)-CRF can achieve the similar recall rate as Electra(base)-CRF through knowledge distillation, while it is 4 times faster. We test three kind of models on a single Tesla P4 GPU in the form of calling flask service, details are shown in Figure 4.
\begin{figure}[h!]
\centering
\includegraphics[scale=0.43]{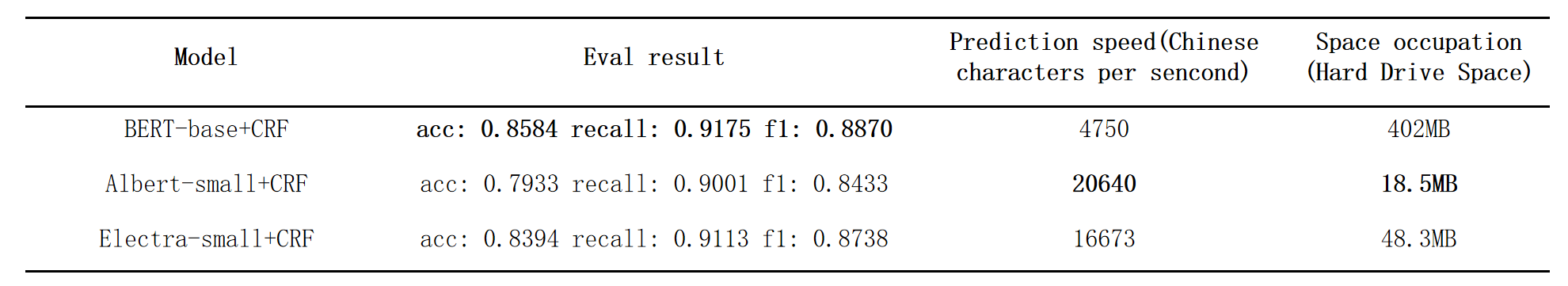}
\caption{Effects of knowledge distillation}
\label{fig:PredictionSpeed}
\end{figure}
\subsection{Workflow}
We represent our working process as in Figure 5.
\begin{figure}[h!]
\centering
\includegraphics[scale=0.5]{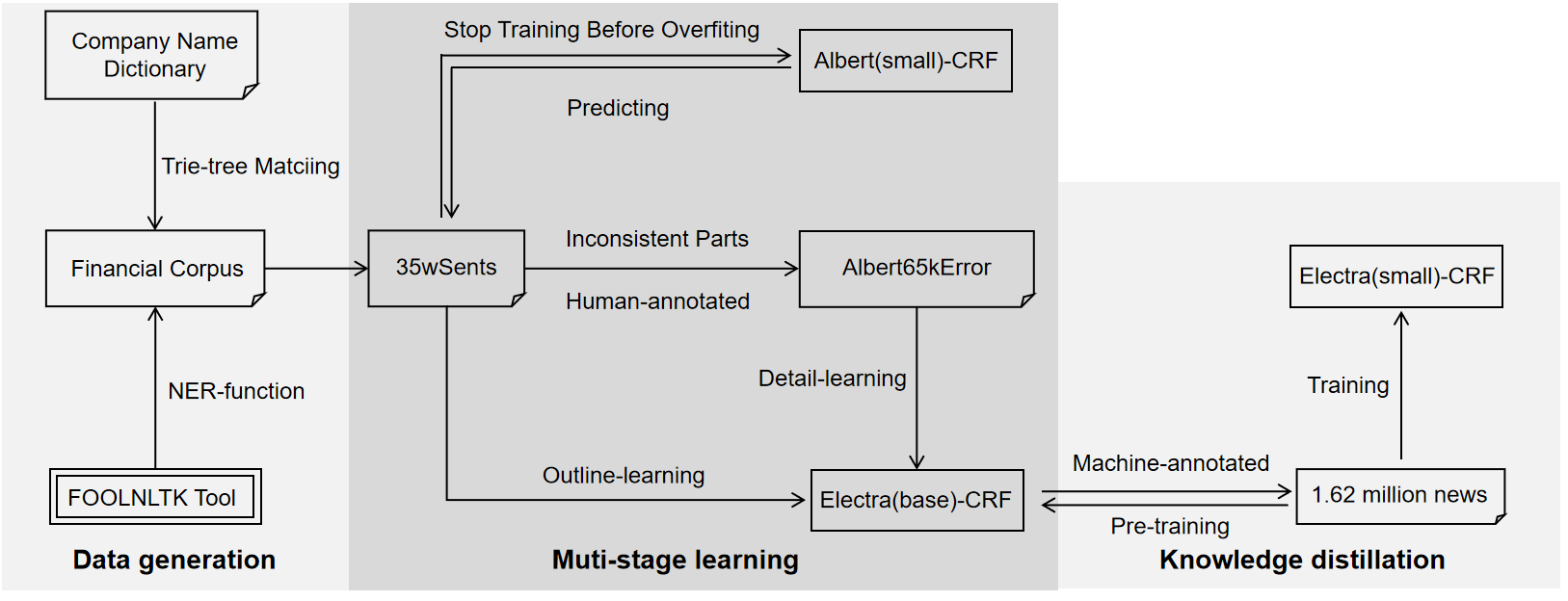}
\caption{framework}
\label{fig:framework}
\end{figure}
\section{DATASET}
In order to improve the performance of the model, we generated a large amount of machine-annotated datasets based on the existing company names dictionary and massive financial news corpus, but we can only share part of them due to business permissions. 
\subsection{Dictionary of full names and abbreviations of listed companies}
Through years of professional work in financial field, we have collected the names of about one million listed company names. With some naming conventions, we can also generate abbreviations for some companies. We use the entire thesaurus as a trie tree matching dictionary.
\subsection{350,000 machine-annotated sentences}
We do trie-tree matching on 350,000 financial sentences with the dictionary of company names. Then we use the NER function of the chinese nlp tool such as FOOLNLTK to recogonize some omissive company names. So we generated a coarse training dataset which is named "35wSents".
\subsection{65,000 manually annotated sentences}
We train a Albert(small)-CRF model on the "35wSents" dataset and stop training before the model get over-fitting.
The obtained model is uesd to predict the "35wSents" dataset, and the sentences which the labels are inconsistent with the prediction are selected for manual re-annotation. We used the community open source YEDDA \citep{Yang2017YEDDA} annotation tool, and we also wrote a format conversion tool to highlight discrepancies between predictions and annotations which reduces the labor costs. The final dataset is named "Albert65kError". 
\subsection{Manually annotated 200 financial News}
Our model is eventually used to do NER task on native financial news, so we created a test dataset and a dev dataset that were completely consistent with the scenario. We selecte 200 financial news that all are longer than 200 chinese words, and divided them into multiple sentences by common Chinese punctuation.The development and test datasets each have 100 news. Both adopt multi-person labeling and cross-validation strategy.
\subsection{1.62 million financial news}
We collected chinese financial news in May 2020 from all web-crawlers, and use it as pre-training corpus. And we train a Electra(base)-CRF model in the multi-stage learning strategy, then use it to do predictions on the 1.62 million financial news. The predicted corpus are fed to Electra(small)-CRF model as training dataset.
\section{Conclusion}
We tune the NER algorithm in the financial scenario through using triE tree matching, financial corpus pre-training, multi-stage learning ploicy, knowledge distillation and other methods to improve the model effect. With as little labor cost as possible, we make full use of available resources, including a millions of company names dictionary, the open source Chinese NLP tools, vast financial news, and the state-of-the-art language modeling such as Albert and Electra. Finally, we found a general method to make the small model reach the accuracy of the large model with a small amount of low-quality training data.

\bibliographystyle{plain}
\bibliography{paper}
\end{document}